\documentclass{article}
\usepackage{ijcai22}

\usepackage{times}
\usepackage{soul}
\usepackage{url}
\usepackage[hidelinks]{hyperref}
\usepackage[utf8]{inputenc}
\usepackage[small]{caption}
\usepackage{graphicx}
\usepackage{amsmath,amssymb,amsfonts}
\usepackage{amsthm}
\usepackage{booktabs}
\usepackage{algorithm}
\usepackage{algorithmic}
\usepackage[T1]{fontenc}    
\usepackage{nicefrac}       
\usepackage{multirow} 
\usepackage{microtype}      
\usepackage{caption} 
\usepackage{xargs}
\usepackage[capitalise]{cleveref}
\usepackage[pdftex,dvipsnames]{xcolor}  
\usepackage[disable,colorinlistoftodos,prependcaption]{todonotes}
\def\Mparam{\theta}  
 
\def\Dparam{\upsilon}

\def\comment#1{}

\def\eqref#1{(\ref{#1})}
\def\Beq#1\Eeq{\begin{equation}#1\end{equation}}
\def\Beqo#1\Eeqo{\begin{equation*}#1\end{equation*}}
\def\Beqs#1\Eeqs{\begin{align}#1\end{align}}
\def\Beqso#1\Eeqso{\begin{align*}#1\end{align*}}

\newtheorem{thm}{Theorem}
\newtheorem{lemma}[thm]{Lemma}
\newcommand\ours{{LogCL}}
\title{Logarithmic Continual Learning} 

\iftrue

\author{
Wojciech Masarczyk$^1$\footnote{Contact Author}\and
Paweł Wawrzyński$^1$\and
Daniel Marczak$^1$\and
Kamil Deja$^1$\And \\
Tomasz Trzciński$^{1,2,3}$\\
\affiliations
$^1$Warsaw University of Technology\\
$^2$Tooploox\\
$^3$Jagiellonian University\\
\emails
wojciech.masarczyk@gmail.com,\\
\{pawel.wawrzynski, kamil.deja, tomasz.trzcinski\}@pw.edu.pl
}

\fi

\usepackage{natbib}

\begin{document}

\maketitle              

\begin{abstract}        
We introduce a neural network architecture that logarithmically reduces the number of self-rehearsal steps in the generative rehearsal of continually learned models. In continual learning (CL), training samples come in subsequent tasks, and the trained model can access only a single task at a time. To replay previous samples, contemporary CL methods bootstrap generative models and train them recursively with a combination of current and regenerated past data. This recurrence leads to superfluous computations as the same past samples are regenerated after each task, and the reconstruction quality successively degrades. In this work, we address these limitations and propose a new generative rehearsal architecture that requires at most logarithmic number of retraining for each sample. Our approach leverages allocation of past data in a~set of generative models such that most of them do not require retraining after a~task. The experimental evaluation of our logarithmic continual learning approach shows the superiority of our method with respect to the state-of-the-art generative rehearsal methods. 

\end{abstract}

\begin{figure}[!t]
\includegraphics[width=0.5\textwidth]{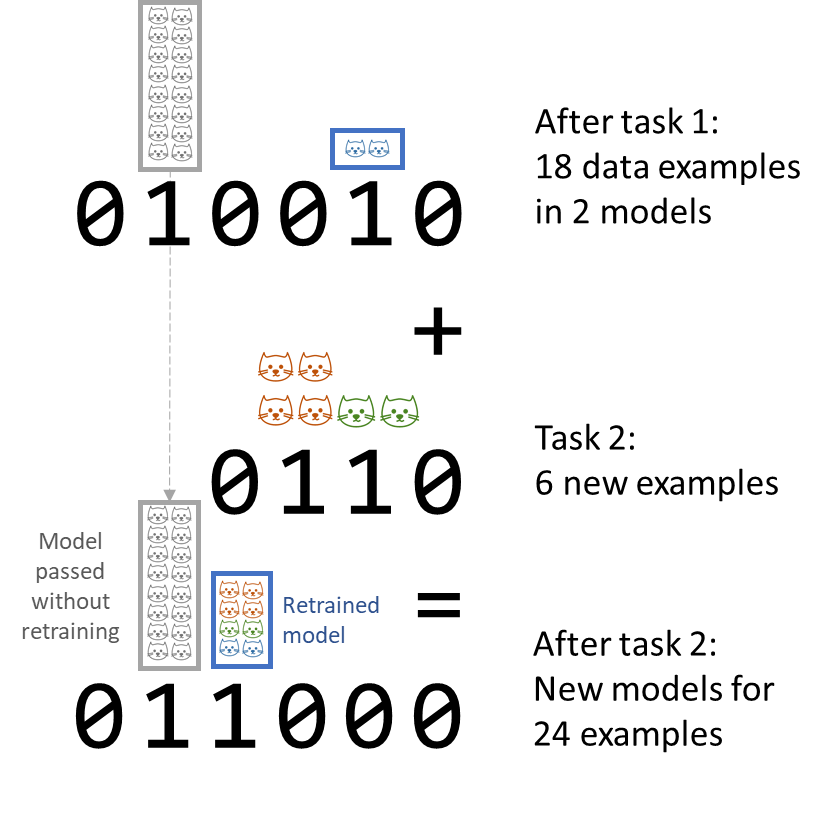}
\caption{
Cats are lazy and don't like to walk too much. Therefore we designed a method that moves as few cats as possible. 
When new data arrives a set of local models is rearranged so $i$-th model is responsible for $2^i$ samples. This procedure is inspired by addition of numbers in a binary system where 1 corresponds to a model and 0 to an empty slot. The most sizeable models are rarely rearranged and, hence, retrained. That leads to slower deterioration of samples reconstructions.} 
\label{fig:teaser} 
\end{figure} 

\section{Introduction} 
\begin{figure*}[!t]
    \centering
    \includegraphics[width=\textwidth]{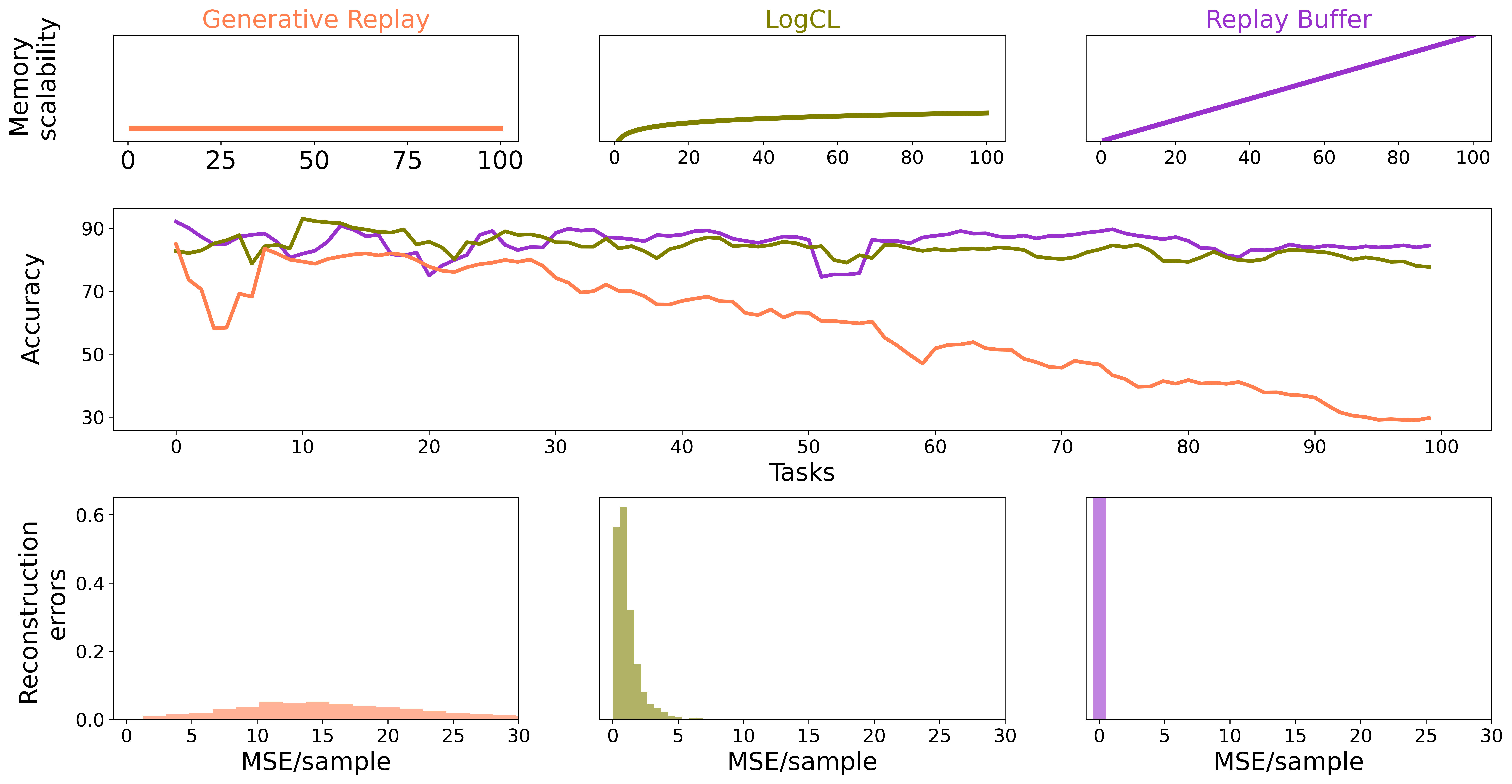}
    \caption{The results of our Extreme Continual Learning experiment (see Sec.~\ref{sec:extreme}). We compare the performance of Generative Replay, LogCL, and Replay Buffer on classifying a sequence of 100 tasks. The first row shows the scalability of memory needed for particular methods. The middle row presents the accuracy obtained by the classifier trained on the data provided b by these methods. The bottom row presents the normalized histogram of reconstruction errors obtained by the respective methods. Note that Replay Buffer achieves 0 error across all data samples since it directly stores the original samples.}
    \label{fig:toy_experiment}
\end{figure*}

Across many applications of neural networks, such as predictive maintenance or surveillance video analysis~\citep{doshi2020continual}
data used for training comes in subsequent tasks. 
On the other hand, the model is expected to adjust its responses to the new data without losing past examples.
Unfortunately, straightforward model retraining leads to {\it catastrophic forgetting} defined as an abrupt performance loss on previously learned skills when acquiring new knowledge~\citep{1999french}.

\emph{Continual learning} (CL) is a machine learning domain that aims to mitigate catastrophic forgetting and enable models to be trained with an incoming stream of training data. This is usually achieved through regularization~\citep{2017kirkpatrick+many}, adaptation of model's architecture~\citep{2016rusu+7} or replay of previous data examples.
Typically, methods based on replay buffer achieve the best performance due to the high quality of stored data samples. However, the memory consumption of these methods grows linearly with the number of tasks which, despite their promising performance, makes them poor candidates to solve the problem of Continual Learning. As shown in~\citep{prabhu2020gdumb}, replay buffer must contain at least a few representatives per class to achieve reasonable performance. Therefore replay buffers with fixed memory budgets are not a feasible solution to the problem of learning from a continuous stream of data.

The approach that tries to address this problem is generative rehearsal~\citep{2017shin+3,van2018generative,von2019continual,2021deja+4} which utilizes a single generative model to rehearse past data and retrain a model with a combination of current and regenerated samples. Memory consumption of such a method is fixed throughout the whole training. However, the main limitation of those methods is that the quality of regenerated samples degrades over time. 

In this paper, we propose a method that finds the sweet spot between naive generative replay and buffer-based methods with a balance between the quality of generated samples and memory consumption. Our method, called Logarithmic Continual Learning (\ours{}),
allocates incoming data samples between generative models of restricted capacity as depicted in Fig~\ref{fig:teaser}. In our approach, the $i$-th model stores exactly $2^i$ samples. Therefore the total number of utilized models equals at most $\log_2 n$, where $n$ is a number of data samples seen so far. When a new task arrives, we gather samples from lower index models and reallocate them to the higher ones according to their capacity. This procedure creates space for new data in models with lower indices and introduces a natural order -- the bigger the model, the older the samples it stores, and the less frequent it is retrained. Samples occupying models that do not need to be repacked remain intact. The allocation strategy, which is inspired by binary numbers addition, limits self-rehearsal to just a small part of all stored samples as usually only part of the bits flip during binary addition. Moreover, assuming that the size of the task is roughly the same in all tasks, it takes bigger models exponentially more time to repack again. Thus the older the sample, the slower it degrades.

Thanks to this design, our method is characterized by these properties:
\begin{itemize}
    \item The maximum number of models scales logarithmically with respect to the number of samples. 
    \item In the worst case, the number of a given sample retraining scales logarithmically with respect to a number of passed tasks.
    \item At each task, only a small subset of models are retrained. The rest remains untouched.
\end{itemize}

To summarize, the main contribution of this work is the introduction of a novel generative rehearsal architecture LogCL that allocates samples to-be-regenerated across $\log_2 n$ generative models to reduce the number of past data reconstructions while maintaining the quality of reconstructed samples.
Last but not least, we provide an extensive evaluation of \ours{} that confirms its superior performance over the competing state-of-the-art methods on a diversified set of benchmark datasets.

\section{Related work} \label{sec:related work} 

There are three main approaches for the continual learning of neural models.

\paragraph{Methods based on 
regularization.}\!\!\!\! The common idea of these methods is to train a~model on the subsequent tasks while regularizing them to preserve the model's performance on the previous data. This can be done by slowing down the learning of model weights selected as significant for previous tasks. Particular methods such as SI~\citep{2017zenke+2} and EWC~\citep{2017kirkpatrick+many} vary on the regularization functions. 
In several works such as NCL~,\citep{kao2021natural} authors combine weights regularization with other techniques, in this case, gradient projection~\citep{saha2021gradient}.

\paragraph{Methods based on task-specific model components.}
These methods build structurally different model versions for different tasks. The sample is first assigned to the proper task at inference time, and the corresponding model version is used. In (PNN) \citep{2016rusu+7}, (DEN) \citep{2017yoon+3}, and (RCL) \citep{2018xu+1} new structural elements are added to the model for each new task, while in \citep{2018masse+2,2019golkar+2,2020wortsman+6} a~large model is considered from which submodels are selected for subsequent tasks. Methods in this category exhibit high accuracy in a task incremental scenario when test samples are given with a corresponding task index~\citep{vandeven2019scenarios}. Otherwise, it has to be assigned through heuristics to solve a crucial part of the continual learning problem~\citep{masana2020class}.

\paragraph{Methods based on replaying.}\!\!\!\! 
Methods in this group are based on the similar assumption that instead of preventing forgetting, we can rehearse previous memories through some form of preserved previous data. On a~new task, the model is retrained with it along with the previous, restored data. Methods presented in \citep{rebuffi2017icarl,2019rolnick+4,hayes2019memory} employ a~memory buffer to store a sample of data examples. \cite{aljundi2019gradient} propose to select memory samples that diversifies gradient directions, while MIR~\citep{aljundi2019online} chooses them according to the changes in the loss function. Recent examples of those methods include GMED by \cite{jin2021gradient}, where stored examples in the buffer are altered to be more challenging for future replay. An interesting approach was proposed by~\cite{prabhu2020gdumb}, where authors use a greedy algorithm for selecting examples for the buffer and retrain the model from scratch with only buffered data. This surprisingly outperforms many recent continual-learning solutions
Nevertheless, storing examples from each task requires a growing buffer what makes this solution inadequate for the general continual learning problem, in which we would like to retrain the model potentially infinitely.

Therefore, \citet{2017shin+3} propose to replace the buffer with a~generative model in the form of the Generative Adversarial Network (GAN) \citep{2014goodfellow+7}. Any structure used to regenerate past data may also suffer from catastrophic forgetting. To avoid it, the authors propose a self-rehearsal procedure to train the generative model with both new data and regenerations of previous examples. \citet{van2018generative} extend this idea to Variational Autoencoder (VAE) by \citet{kingma2014autoencoding}. Additionally, the authors combine the generative model with the base classifier, which reduces the cost of model retraining. 

An interesting idea that we can place in between generative and buffer-based rehearsal is presented by \citet{caccia2020online} who incorporate VQ-VAE~\citep{oord2018neural} architecture to compress the original data examples into a~special representation that requires less memory than original images. 

In HyperCL~\citep{von2019continual}, the authors introduce a general approach where the weights of continually trained networks are generated by another model called hypernetwork. In this work, the authors propose two approaches, one with explicit classifier generation from a hypernetwork and the second one in which task-independent generative models are created to generate rehearsal samples for the classifier.

\citet{2021deja+4} proposed BinPlay, an architecture for generative rehearsal continual learning. BinPlay, rather than being trained to generate data similar to the observed samples, learns to reconstruct only the observed examples. 

However, when subsequent tasks occur, the model is trained in generative rehearsal what inevitably leads to the degradation of previous generations. 

In this work, we propose a method that balances the need for maintaining high-quality samples and low memory consumption.

\section{Problem definition} 
\label{sec:problem} 

We analyze the most typical scenario in which the need of CL occurs. The data comes in input-output pairs 
\Beq \label{in-out:pair} 
    \langle x_i, y_i \rangle, i=1,2,\dots 
\Eeq 

where $x_i$ denotes images and $y_i$ labels.

The data comes in subsequent tasks. The last sample index in $n$-th task is $i_n$. When $n$-th task is available, no other data is directly accessible. The tasks are generally of different sizes. 

The goal is to have a model, $f(x;\Mparam)$, with weights $\Mparam$. After accessing the $n$-th task, the model minimizes the average loss

\Beq \label{quality index} 
    \frac1{i_n} \sum_{i=1}^{i_n} L(y_i, f(x_i; \Mparam)), 
\Eeq 
where $L$ is a certain loss function, such as the squared Euclidean distance or CrossEntropy. Note that $x_i$ may as well contain a code of the index to formulate a Task Incremental scenario, however in our experimental study we only consider a harder Class Incremental scenario, where $x_i$ is a~vector of raw data.

\section{Method}
\label{sec:method} 
\begin{figure*}[t]
\centering
\includegraphics[width=0.8\textwidth]{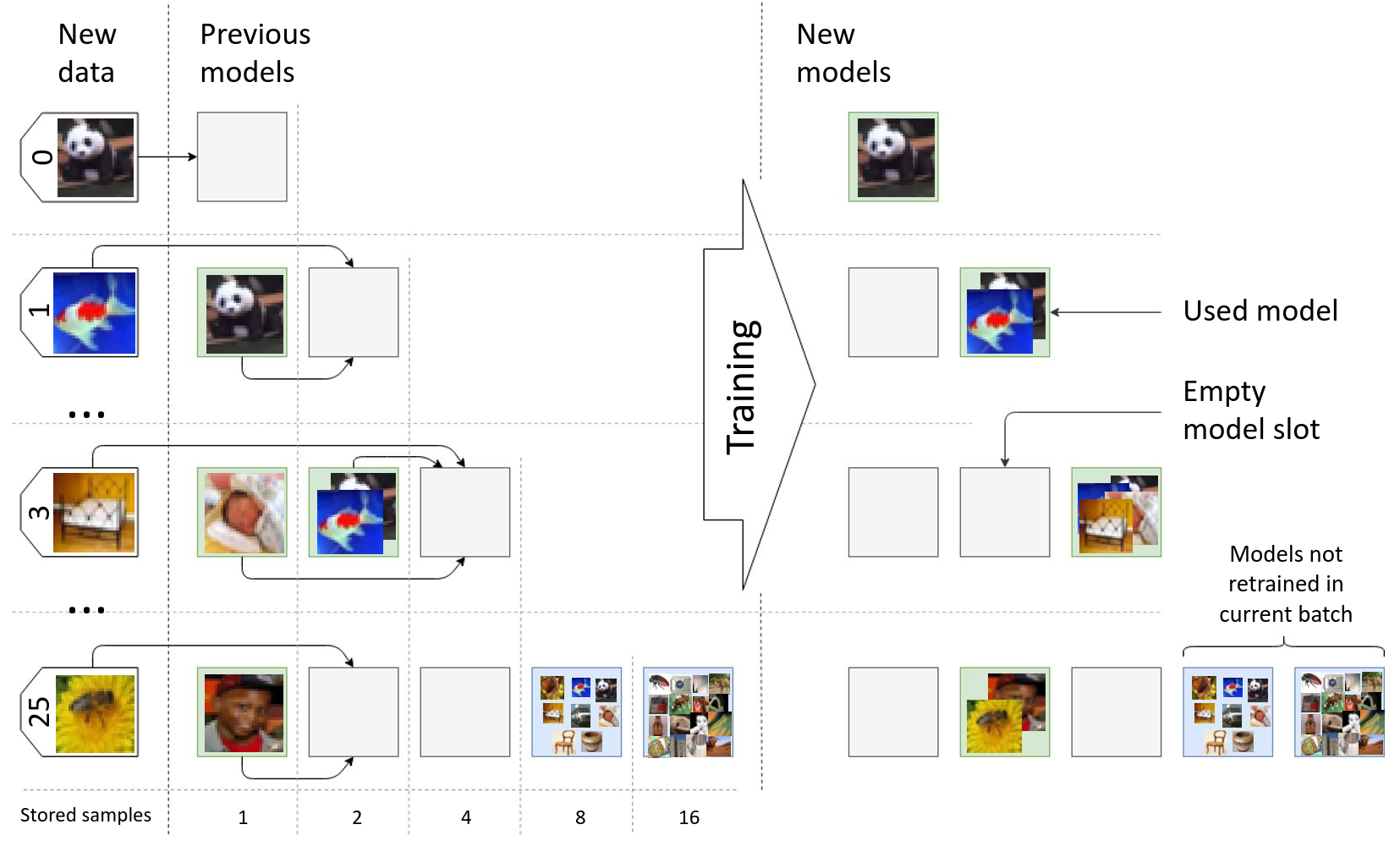}
\caption{Overview of the method. When new data arrives, a set of local models is rearranged, so the $i$-th model is responsible for $2^i$ samples. This procedure is inspired by adding numbers in a binary system where one corresponds to a model and 0 to an empty slot. The most sizeable models are rarely rearranged and, hence, retrained. That leads to slower deterioration of samples.} 
\label{fig:method_overview} 
\end{figure*}

In this section, we introduce how our LogCL works. The primary goal of our method is to enable the storage of data from past tasks in a set of binary autoencoders  minimizing both degradations of samples and the memory footprint of the method. Therefore we propose a solution for deploying incoming samples inspired by binary coding.  In our solution, these autoencoders correspond to 1-s in the binary notation of the total number of samples arrived so far. We first rearrange already encoded examples with each new task to prepare a space for incoming ones. To that end, we associate each image index with a binary code that directly defines in which model the image should be stored. This indexing function assigns recent images into models that are not yet fully occupied and contain relatively fresh examples. This approach minimizes the number of times an image has to be trained based on its generation, thereby accumulating the reconstruction error. Fig.~\ref{fig:method_overview} provides an overview of our method, and below, we describe it in more detail, along with the model properties and extensions, which motivate the design choices we make.

\subsection{Architecture}

The whole architecture is based on a~local part and a~global part. The {\bf local part} is composed of pairs. The $k$-th pair, $k=0,1,\dots$, contains:  
\begin{itemize} 
\item 
Image decoder, $d(c;\Dparam_k)$, with weights $\Dparam_k$. It translates codes, $c$, of image indices into the images. Each local model operates on a~different pool of images and their codes. 
\item 
Mapping, $f(x;\Mparam_k)$, with weights $\Mparam_k$. It translates images, $x$, into their labels. 
\end{itemize} 
The {\bf global part} is a mapping, $f(x;\Mparam^g)$, that translates images, $x$, into their labels using weights $\Mparam^g$. 

\subsection{Assignment of samples to local models} 

Assignment of samples, indexed by $i$, to local models, indexed by $k$, is based on the following rules: 
\begin{description} 
\item[R1:] 
After the $n$-th task, all $i_n$ samples of past data are reassigned to the local models. 
\item[R2:] 
To $k$-th local model, either $2^k$ data samples are assigned, or none.
\item[R3:] 
The assignment of sample indices to local model indices, $i\mapsto k$, is nonincreasing. 
\end{description} 

Consequently, after each $n$-th task, the assignment of samples to local models can easily be designated based on the binary notation of $i_n$. For instance, let $i_n=10=\verb'b1010'$. Then, we have 
\begin{itemize} 
\item 3-rd local model is trained with $2^3$ samples, namely $\{1,2,3,4,5,6,7,8\}$. 
\item 2-nd local model is not used. 
\item 1-st local model is trained with $2^1$ samples, namely $\{9,10\}$. 
\item 0-th local model is not used. 
\end{itemize} 

\subsection{Training}

On a new task, some local models (usually very few of them) are trained, and the global model is trained. The training aims to achieve two goals: First, we will satisfy rules R1-R3 of sample assignment to local models. Secondly, the global model needs to be effectively trained on samples of the same quality, even though they are unequally distorted by replaying them from local models and using them to train others. If this issue is not addressed, then the global model is likely to infer labels of images based on their quality. 

To determine which local model should be trained, we look at the binary notation of a~total number of samples before the task and the total number of samples after. Let these numbers be $i_n$ and $i_{n+1}$, respectively. We look at the most significant bit that differs in $i_n$ and $i_{n+1}$. The local models that correspond to that bit and the less significant bits need to be retrained with the new data and the data regenerated from the ``old'' local models. 

For instance, let $i_n=10=\verb'b1010'$ and a~new task has arrived with $3$~samples. Their indices are $\{10,11,12\}$. Now $i_{n+1}=13=\verb'b1101'$. $2$-nd bit is the most significant that differs in $i_n$ and $i_{n+1}$. Now we have: 
\begin{itemize} 
\item 3-rd local model remains unchanged. 
\item 2-nd local model needs to be trained with the samples $\{9,10,11,12\}$. Among them, samples $\{9,10\}$ are replayed from the previous 1-st model, and samples $\{11,12\}$ are new. 
\item 1-st local model is now not used. 
\item 0-th local model is trained with the new sample $\{13\}$. 
\end{itemize} 

With each task the global model is retrained. Let the new total number of samples be $i_{n+1}$. The retraining is based on sampling image-label pairs from the local models and using them to train the global model. The retraining consists of repeating the following: 
\begin{enumerate} 
\item 
$i \sim U(\{1,\dots,i_{n+1}\})$, a~sample index.  
\item 
$k$ is determined such that $k$-th local model has been trained with the use of $i$-th sample (among others). 
\item 
$x \leftarrow d(c(i);\Dparam_k)$.   
\item 
$y \leftarrow f(x;\Mparam_k)$. 
\item 
The global model is trained with the pair $\langle \text{aug}(x),y\rangle$, where aug is a~method of data augmentation.
\end{enumerate}
The augmentation in point 5. addresses the following unwanted effect: Images generated by a~local decoder may have specific common distortions; the global model may use these distortions to assign labels to the images. An~augmentation in the form of white noise added to the images prevents this effect. 

\subsection{Properties} 
\label{sec:properties} 

Thanks to the design choices inspired by binary notation, our method features the following desired properties:
\begin{enumerate}
    \item The maximum number of local models required for total $i_n$ data samples is $\lceil \log_2 i_n\rceil+1$. 
    \item The average number of local models to store $i_n$ data samples equals $\frac{\lceil \log_2 i_n\rceil+1}{2}$.  
    \item Any data samples undergoes the retraining phase at most $\lceil \log_2 i_n\rceil$ times. (see Lemma 1)
    \item The gaps between retraining particular samples grow exponentially
    \item Usually only models with small indices and little reconstructed data are rearranged
\end{enumerate}

The property of maximum number of models to store $i_n$ samples follows from the fact that the joint capacity of these models can be written as: $2^0+2^1+\dots+2^{\lceil \log_2 i_n\rceil}=2^{\lceil \log_2 i_n\rceil+1}-1 \geq i_n$.

The second property follows from the fact that, on average, to represent $n$-bit number in binary coding, we use $\frac{n}{2}$ ones.

Usually, when a~data sample is replayed and used to train another local model, its reconstruction deteriorates. However, that happens fairly few times in our proposed architecture, as the lemma below specifies. 

\begin{lemma} \label{lemma:log} 
Let the presented architecture have been trained with $n$ tasks. Then, any data sample has been replayed and taken part in training another local model at most $\lceil \log_2 i_n \rceil$ times. 
\end{lemma} 
{\it Proof:} The claim is proved in two points. Firstly, whenever a piece of data is read from one source, local model and applied to train another destination, local model, the source model has a lower index than the destination one. Secondly, the highest index of a~local model is $\lceil \log_2 i_n \rceil-1$. 

To prove the first point, we consider the arrival of a new task and training a~local model of index $k$; it is trained with $2^k$ samples. This model can be trained entirely on new data. Otherwise, it is trained with data contained in models with lower indices. There are at most 
$$
    1+ \dots +2^{k-1} = 2^k-1
$$
Such data samples; the rest of the $2^k$ samples required to train the model are new. Therefore, whenever the destination model is trained with data previously contained in the models with lower indices, it sweeps all the data from them. Hence, source models always have smaller indices than the destination ones. 

The second point is an immediate result of how natural numbers are written in binary notation. The index of the most significant bit of the $i_n$ number in $0$-based indexing is $\lceil \log_2 i_n \rceil-1$. 
$\blacksquare$

Also, in a~typical scenario in which tasks are of similar size, only models with small indices and little data are usually rearranged. This is because only its least significant digits change when an~integer is incremented by a~small number. 

\subsection{Decreasing memory footprint} 

\paragraph{Diverse architecture of local models.} 
Let $K$ be the current largest index of a~local model. The $K$-th model stores over half of all the data. A~$j$-th local model ($j<K$) stores $2^{K-j}$ times fewer data.  To save the memory footprint of the whole architecture, we propose to construct local models of diverse sizes: The size of the largest model is fixed, and the size of $j$-th model decreases geometrically with $j$. This way, the memory footprint of all $K+1$ local models could be proportional to the size of the largest local model rather than to this size times $(K+1)$. We adopt this approach in our experimental study, where the ratio of $j$-th model to the size of $K$-th model is $(7/8)2^{j-K}+(1/8)$.

\paragraph{Local models for large volumes of data.} Storing small portions of data in local models is inefficient when these models have a memory footprint larger than the data. 
Therefore, we propose to store $L2^k$ data samples in $k$-th local model, for a~certain $L\in\mathbb{N}$, and keep a~memory buffer of size $L-1$, similarly to \citep{dushyant2019curl}. Details of this solution are presented in the appendix.

\section{Experimental study} 
\label{sec:experiments}
\subsection{Extreme Continual Learning}
\label{sec:extreme}
To show the superiority of our method, we create an unprecedented continual learning experiment with 100 tasks in which the model learns to classify 300 classes from the Omniglot dataset. In this setting, we compare LogCL to the naive generative replay method, where a single model is constantly retrained to store an increasing amount of data within a fixed memory budget. Additionally, we compare LogCL to the replay buffer method, which stores all previously encountered examples in the buffer. To the best of our knowledge, LogCL is the first method that offers sublinear memory scalability and solves the problem of continually learning 100 tasks in a class incremental scenario.

As Fig.~\ref{fig:toy_experiment} shows, our method (olive) consumes significantly less memory than standard replay buffer (violet) yet achieves comparable results in terms of accuracy. In contrast, generative replay (orange) utilizing a single model reconstructs the same data with significantly bigger errors (bottom row), leading to gradual degradation of results over a long sequence of tasks.

\subsection{Main experiments}
In the following sections, we empirically evaluate the performance of our method on three commonly used continual learning benchmarks: MNIST \citep{lecun2010mnist}, Omniglot \citep{lake2015omniglot} and CIFAR-100 \citep{krizhevsky2012cifar}. We use standard splitMNIST with class incremental scenario resulting in 5 tasks. 
To construct a CL task with the Omniglot dataset, we follow the approach of~\citet{dushyant2019curl} and use alphabets as classes, then split the whole dataset into ten tasks of 5 classes each.

Whenever the autoencoder is in use, we apply the architecture introduced by~\citet{2021deja+4} with an adjusted size of the latent space for each dataset~\footnote{Detailed implementation information can be found in the released codebase}.

As a measure of performance, we report the average accuracy of the global classifier on the whole test set after finishing the training sequence. 

\subsection{Results}
The results presented in Table~\ref{tab:results_mnist_omniglot} confirm that our method significantly outperforms other solutions based on generative replay on MNIST and Omniglot. We credit this superior performance to our efficient utilization of multiple generative models, resulting in high-quality reconstructed samples. As Fig.~\ref{fig:toy_experiment} shows, \ours{} offers a much higher quality of consecutive reconstructions than the competing BinPlay~\citep{2021deja+4} which utilizes only a single generative model.

To the best of our knowledge, LogCL is the first generative-based model which achieves a satisfactory performance on the CIFAR-100 dataset in a scenario with 20 splits. Therefore, we compare our method with other methods which utilize replay buffer to tackle the problem of continual learning. To compare our method with these approaches, we take the results from~\citep{MAI202228, prabhu2020gdumb}.

\begin{table}[htb!]
  \centering
  \begin{tabular}{l||cc}
    \toprule
    Model & MNIST & Omniglot\\
    \midrule
    HyperCL & $95.3 \pm 0.37$ & 20.7 $\pm$ 0.37 \\
    \midrule
    VCL & $97.7 \pm 0.05 $ & 13.4 $\pm$ 1.8 \\
    \midrule
    CURL&  $94.2 \pm 0.87 $ & $21.2 \pm 0.74$ \\
    \midrule
    BinPlay &  $97.2 \pm 0.6$ & $38.8 \pm 5.9$ \\
    \midrule
    \textbf{\ours{} (ours) }& \textbf{98.65 $\pm$ 0.09} & \textbf{55.2 $\pm$ 1.15} \\
    \bottomrule
  \end{tabular}
    \caption{Average accuracy after the final task in the class incremental scenario (in \% $\pm$ SEM). Our approach clearly outperforms competitive methods on presented benchmarks.}
  \label{tab:results_mnist_omniglot}
\end{table}

\begin{table}[htb!]
  \centering
  \begin{tabular}{l||cc}
    \toprule
    Model & Accuracy \\
    \midrule
    EWC~\citep{2017kirkpatrick+many} &$9.5 \pm 0.83$\\
    \midrule
    SI~\citep{2017zenke+2} &$13.3 \pm 1.14$\\
    \midrule
    HyperCL~\citep{oswald2020hypercl} & 8.8 \\
    \midrule
    ER~\citep{hayes2019memory} &$18.4 \pm 1.4$\\
    \midrule
    MIR~\citep{aljundi2019online} &$19.3 \pm 0.7$\\
    \midrule
    GSS~\citep{aljundi2019gradient} &$13.4 \pm 0.6$\\
    \midrule
    iCaRL~\citep{rebuffi2017icarl} &$42.9 \pm 0.8$\\
    \midrule
    GDumb~\citep{prabhu2020gdumb} &$28.8 \pm 0.9$\\
    \midrule
    GMED~\citep{jin2021gradient} &$21.22 \pm 1.0$\\
    \midrule
    \textbf{\ours{} (ours) }& \textbf{56.75 $\pm$ 1.0} \\
    \bottomrule
  \end{tabular}
    \caption{Average accuracy after the final task in the class incremental scenario (in \% $\pm$ SEM) for CIFAR-100 dataset with 20 splits. Our approach clearly outperforms competitive methods on the presented benchmark. The results are taken from~\citep{MAI202228,prabhu2020gdumb} or original works.}
  \label{tab:CIFAR-100}
\end{table}

\section{Discussion} 
\label{sec:discussion} 

Our approach is based on generative models that reproduce random previously seen data samples. We hypothesize that an a~properly designed neural network can be a~scalable storage of large volumes of data samples. In this order, these samples need to share some common features. Otherwise, this kind of storage will not be scalable, and storing the data compressed with some typical methods will be a~better choice. 

Further development of the approach presented here will be based on three points. Firstly, its efficiency could be strongly improved by introducing of selection of data samples actually stored in the generative models. Secondly, these generative models could be improved to increase their accuracy and reduce their size. Thirdly, they could also be initially small but enhanced with techniques that trade their compactness for accuracy.

\section{Conclusions} 
\label{sec:conclusions} 

In this paper, we propose an architecture for generative rehearsal continual learning in which the generative models employed are trained with the data they generate alone at least $\log_2 i_n$ times where $i_n$ is the total number of data samples. This comes at the cost of the need for a logarithmic number of instances of this generative model. In the experimental study, we apply our approach to obtain state-of-the-art performance of continual learning on MNIST, Omniglot, and CIFAR-100 datasets.

\bibliographystyle{named}
\bibliography{references}

\end{document}


\newcommand\ours{{LogCL}}

\title{Logarithmic Continual Learning -- appendix} 

\author{%
  First Author, Second Author, Third Author \\
  Institute of Computer Science\\
  Warsaw University of Technology\\
  Nowowiejska 15/19, 00-665 Warsaw, Poland \\
  \texttt{\{firt.author|second.author|third.author\}@pw.edu.pl} 
}

\maketitle              

\begin{abstract}        

This is the supplementary material for our Logarithmic Continual Learning submission. In here we describe detailed training hyperparameters, as well as extended visualization of generations produced by our Logarithmic Continual Learning model in comparison to BinPlay.

\end{abstract}

\section{Training hyperparameters}
We use the following architectures for the the datasets:
\begin{itemize}
    \item SplitMNIST -- the autoencoder has the latent space of size 200. For the classifier, we use a model based on LeNet with additional dropout layers.
    \item SplitOmniglot -- the autoencoder is the same as in the case of SplitMNIST experiments. For the classifier, we use standard ResNet18.
    \item SplitCIFAR-100 -- the autoencoder has the latent space of size 1000. For the classifier, we use standard ResNet18.
\end{itemize}

In case of our Continual Learning in the limits experiment, we used the same hyperparameters as in the Omniglot dataset.

The following section describes hyperparameters used in experimental studies of \ours{}.

We trained all the models with Adam optimizer. For generative models, learning rate was scheduled exponentially with a~scheduler rate equal to $0.99$. The hyperparameters that varied across different scenarios are presented in Tab.~\ref{tab:hyperparameters}.

\begin{table*}[htb!]
  \centering
  \begin{tabular}{c|l||c|c|c}
    \toprule
    & Hyperparameter & MNIST & Omniglot & CIFAR--100\\
    \midrule
    \midrule
    \multirow{5}{*}{\rotatebox[origin=c]{90}{\parbox[t]{1cm}{\centering Classifier}}} & Batch size & 120 & 40 & 128 \\
    \cmidrule{2-5}
    & Learning rate & 0.002 & 0.001 & 0.1\\
    \cmidrule{2-5}
    & Epochs & 15 & 10 & 65 \\
    \midrule
    \midrule

    \multirow{8}{*}{\rotatebox[origin=c]{90}{\parbox[t]{1cm}{\centering Generator}}} & Batch size & 40 & 40 & 50 \\
    \cmidrule{2-5}
    & Warm-up learning rate* & 0.001 & 0.0001 & 0.001 \\
    \cmidrule{2-5}
    & Warm-up epochs* & 30 & 50 & 100 \\
    \cmidrule{2-5}
    & Learning rate & 0.01 & 0.05 & 0.005 \\
    \cmidrule{2-5}
    & Epochs & \multicolumn{3}{c}{200} \\
    
    \bottomrule
  \end{tabular}
    \caption{Hyperparameters used in \ours{} experiments. * Warm-up is a phase of generative model training when binary codes are not assigned to particular samples.}
  \label{tab:hyperparameters}
\end{table*}